# Communications to Circulations: Real-Time 3D Wind Field Prediction Using 5G GNSS Signals and Deep Learning

**Yuchen Ye[1,2], Chaoxia Yuan[1,3]*, Mingyu Li[1,2], Aoqi Zhou[1,2], Hong Liang[4,5]*, Chunqing Shang[6], Kezuan Wang[7], Yifeng Zheng[7], Cong Chen[7]**

1 State Key Laboratory of Climate System Prediction and Risk Management (CPRM) /Key Laboratory of Meteorological Disaster, Ministry of Education (KLME) /Collaborative Innovation Center on Forecast and Evaluation of Meteorological Disasters (CIC-FEMD), Nanjing University of Information Science and Technology, Nanjing 210044, China

2 School of Atmospheric Sciences, Nanjing University of Information Science and Technology, Nanjing 210044, China

3 School of Artificial Intelligence, Nanjing University of Information Science and Technology, Nanjing 210044, China

4 State Key Laboratory of Environment Characteristics and Effects for Near-space, Beijing 100081, China

5 Engineering Technology Research Center for Meteorological Observation of CMA, Beijing 100081, China

6 Huawei Technologies Co., Ltd, Guangdong 518129, China

7 China Mobile Zhejiang Co., Ltd, Zhejiang 310009, China

Corresponding author: Chaoxia Yuan (chaoxia.yuan@nuist.edu.cn); Hong Liang

(liangh@cma.gov.cn)

## Abstract

Accurate atmospheric wind field information is crucial for various applications, including weather forecasting, aviation safety, and disaster risk reduction. However, obtaining high spatiotemporal resolution wind data remains challenging due to limitations in traditional in-situ observations and remote sensing techniques, as well as the computational expense and biases of numerical weather prediction (NWP) models. This paper introduces G-WindCast, a novel deep learning framework that leverages signal strength variations from 5G Global Navigation Satellite System (GNSS) signals to forecast three-dimensional (3D) atmospheric wind fields. The framework utilizes Forward Neural Networks (FNN) and Transformer networks to capture complex, nonlinear, and spatiotemporal relationships between GNSS-derived features and wind dynamics. Our preliminary results demonstrate promising accuracy in real-time wind forecasts (up to 30 minutes lead time). The model exhibits robustness across forecast horizons and different pressure levels, and its predictions for wind fields show superior agreement with ground-based radar wind profiler compared to concurrent European Centre for Medium-Range Weather Forecasts (ECMWF) Reanalysis v5 (ERA5). Furthermore, we show that the system can maintain excellent performance for localized forecasting even with a significantly reduced number of GNSS stations (e.g., around 100), highlighting its cost-effectiveness and scalability. This interdisciplinary approach underscores the transformative potential of exploiting non-traditional data sources and deep learning for advanced environmental monitoring and real-time atmospheric applications.

## Introduction

Atmospheric wind fields are a fundamental component of the Earth system. They govern the transport of momentum, heat, moisture, and pollutants [1, 2]. Accurate knowledge of atmospheric wind fields and their evolution is therefore essential for a broad range of applications, including weather forecast and climate prediction, aviation safety, air quality management, renewable energy generation, and disaster risk reduction [3, 4]. Yet, despite their significance, the wind fields remain among the most challenging atmospheric variables to observe and forecast in high spatiotemporal resolution[5-7].

Traditionally, wind information has been obtained through direct in situ observations such as radiosondes, pilot balloons, aircraft-mounted sensors, and ground-based radar wind profiler (RWP) [8-13]. While these measurements offer high vertical or temporal resolution in localized regions, their spatial coverage remains limited, particularly over oceans and remote or data-sparse areas [14-16]. To enhance the global coverage of wind observations, a variety of remote sensing techniques have been developed[17]. Among these, satellite-derived wind prediction methods are most prominent [18-20]. Atmospheric Motion Vectors (AMVs) are obtained by tracking the displacement of cloud or water vapor features in successive geostationary or polar-orbiting satellite images [18, 21, 22]. Ocean surface winds are estimated to be using scatterometer data by relating microwave backscatter to surface roughness[23-25]. Recently, active sensing instruments, such as the Doppler wind lidar onboard the ESA Aeolus satellite, have been used to measure horizontal wind profiles directly from space[26-28]. These remote sensing techniques have substantially improved the spatiotemporal resolution and global consistency of wind field datasets.

Despite these advances, several challenges persist[29, 30]. AMV predicts are sensitive to cloud tracking errors and are limited in vertical resolution[20, 31]. Scatterometer measurements provide only surface winds and are confined to ice-free ocean regions [32]. Doppler lidar instruments are still relatively new, and their coverage and data quality are constrained by instrument design and orbital parameters [33, 34]. Additionally, while numerical weather prediction (NWP) models assimilate observational data to forecast wind fields, they remain computationally expensive and susceptible to initialization uncertainties and model biases, particularly in complex terrain or convective environments[35-37].

In recent years, the increasing availability of unconventional data sources and the rapid development of machine learning techniques have opened new avenues for atmospheric wind estimation[38-42]. In this context, we propose a novel approach to forecast atmospheric wind fields by leveraging signal strength variations in Global Navigation Satellite System (GNSS) communication signals, coupled with deep learning-based modeling. This method exploits the sensitivity of GNSS signals to atmospheric refractivity, which is influenced by meteorological variables such as pressure, temperature, and humidity—all of which are closely linked to wind field variations[43]. The underlying physical mechanism is that GNSS signals experience attenuation, delay, and phase shifts as they propagate through the atmosphere[44]. These perturbations, often treated as sources of error in positioning applications, can be reinterpreted as valuable indicators of atmospheric state[45, 46]. By systematically collecting and processing GNSS signal intensity data from dense ground-based receiver networks, it is possible to extract features indicative of atmospheric dynamics[47].

To capture the complex, nonlinear, and spatiotemporal relationships between GNSS signal variations and wind fields, we adopt deep learning techniques, including Forward neural networks (FNN) and Transformer networks. These architectures are particularly well-suited for high-dimensional data and time series prediction tasks. Through supervised learning on historical datasets that pair GNSS-derived features with wind field observations (e.g., from wind profiling radar), the model can learn to infer current wind states and forecast future wind evolutions.

This approach offers several potential advantages. First, it enables high-resolution wind field estimation. Second, it is cost-effective and scalable, building upon existing GNSS infrastructure without the need for deploying additional physical sensors. Third, the deep learning framework allows for rapid inference, supporting real-time applications such as nowcasting and early warning systems. More broadly, this work represents an interdisciplinary integration of atmospheric science, satellite communications, and artificial intelligence. It exemplifies the transformative potential of data-driven methods in geophysical research and underscores the value of exploiting nontraditional data sources for environmental monitoring. Preliminary results indicate that our GNSS-deep learning system achieves promising accuracy in real-time wind forecasting.

**5G GNSS Signal Wind Field Forecast Deep Learning Model (G-WindCast) Framework**

The schematic diagram of G-WindCast is shown in Fig. 1. The overall workflow of the G-WindCast framework consists of three components: 1) Data preprocessing, which converts GNSS data and wind field data into the model input format (details of the 5G GNSS data and wind field data used can be found in Supplementary Information Section E); 2) Construction of a deep learning model (model details are provided in Supplementary Information Section A; evaluation metrics used are described in Supplementary Information Section C; training details are given in Supplementary Information Section D); and 3) Post-processing of the output wind field (details are provided in Supplementary Information Section B). The data preprocessing module primarily removes outliers and interpolates missing values while aligning the temporal resolution of GNSS data and wind field data. The deep learning model employs neural networks to predict wind fields using GNSS data as input, identifying statistical relationships between GNSS signals and wind fields. The post-processing step adjusts the mean and variance of the predicted wind field to better align with the true data distribution of the training set. In experiments, our prediction time steps range from a 5-minute lead to a 30-minute lead to evaluate the sustained performance of our deep learning model. For performance comparison, we benchmarked against ERA5—one of the state-of-the-art reanalysis datasets—by comparing our model's 30-minute predictions and ERA5 wind field data at the same time points against the ground-based RWP. The model was evaluated both regionally (multi-station) and at single stations. In single-station experiments, we used only limited GNSS data to derive point-specific winds, assessing model performance under severely data-deficient scenarios (e.g., single-airport deployments).

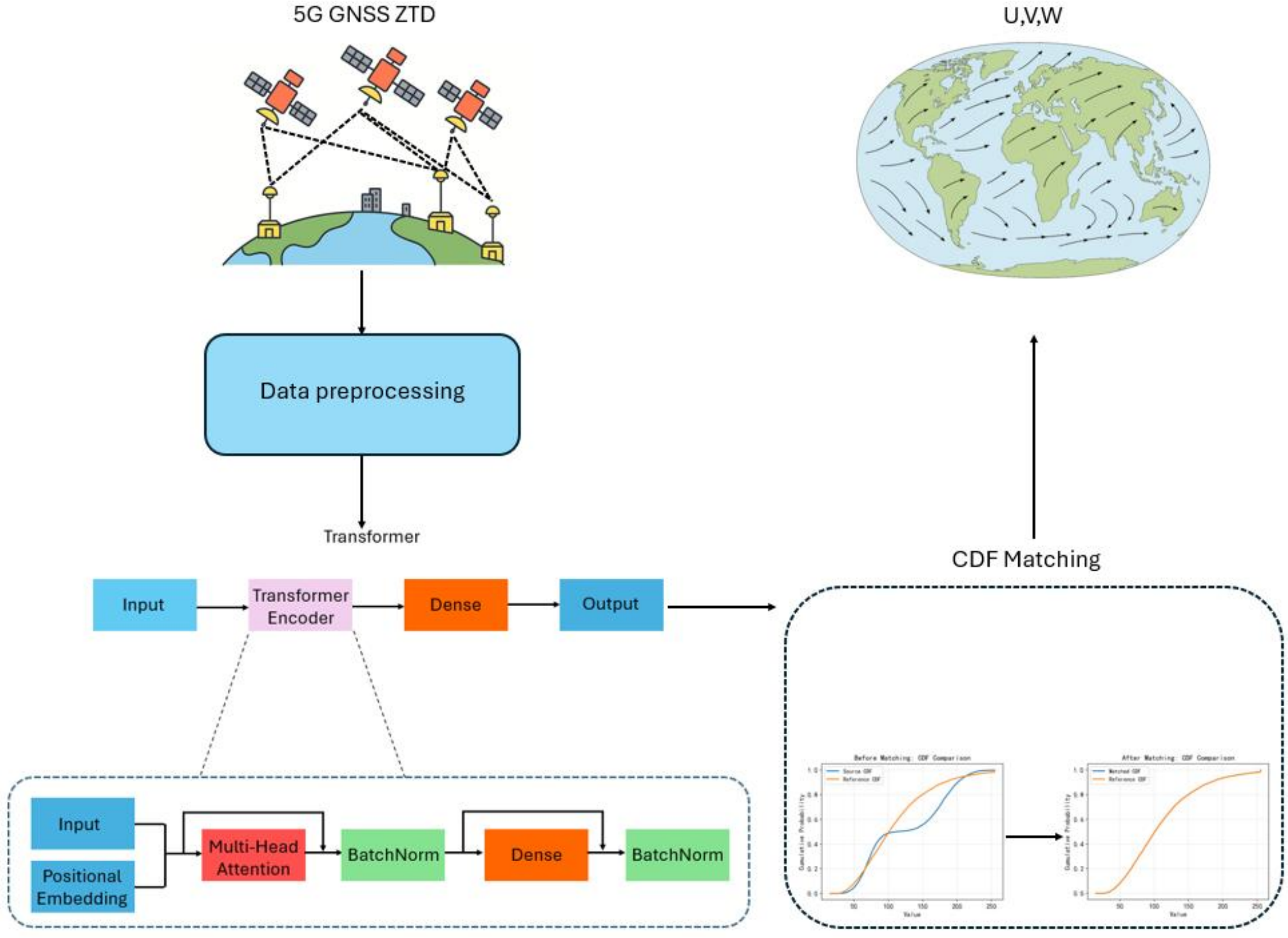

**Fig. 1 | 5G GNSS signal wind field forecast deep learning model (G-WindCast) framework**

## Region-mean performance of G-WindCast

Fig. 2 illustrates the performance of G-WindCast, presenting a mosaic plots based on different forecast lead times (horizontal axis, 5 to 30 minutes) and varying pressure levels (vertical axis, 1000 to 400 hPa) averaged over the region of interest, Zhejiang province of China. For RMSE, MAE, and RMSPE, lower values (white) indicate better performance, while higher values (red) reflect greater errors. From Fig. 2, it can be observed that for forecast lead times ranging from 5 to 30 minutes, the metrics for U (zonal wind), V (meridional wind), and W (vertical wind) show no significant variation, indicating that our model maintains robust performance without notable degradation within this time frame. The RMSE and MAE for U and V are larger at higher altitudes, which is related to their magnitude differences across pressure levels—U and V typically exhibit smaller magnitudes at lower levels and larger magnitudes at higher levels, leading to higher RMSE and MAE in the upper atmosphere[48]. In contrast, W behaves differently because it is generally weaker and more influenced by surface friction and local turbulence near the ground, making it more challenging to predict[49]. Our model currently does not incorporate any surface-related predictors (e.g., evaporation, precipitation, surface roughness), but future experiments could explore adding such factors to improve near-surface prediction accuracy. Regarding RMSPE, U and V show relatively uniform distributions across pressure levels, whereas W exhibits a high-value region at 400 hPa. We attribute this likely to the data's inability to capture the gradient of W between 400 hPa and other pressure levels—there is no predictive output above 400 hPa, and the distance to the 300 hPa level is substantial (as the geopotential height difference per 100 hPa increases with altitude). Overall, our model demonstrates strong performance for U, V, and W, with RMSPE consistently below 0.04 (4%), indicating high accuracy. Additionally, the model does not exhibit significant performance degradation with increasing forecast lead times.

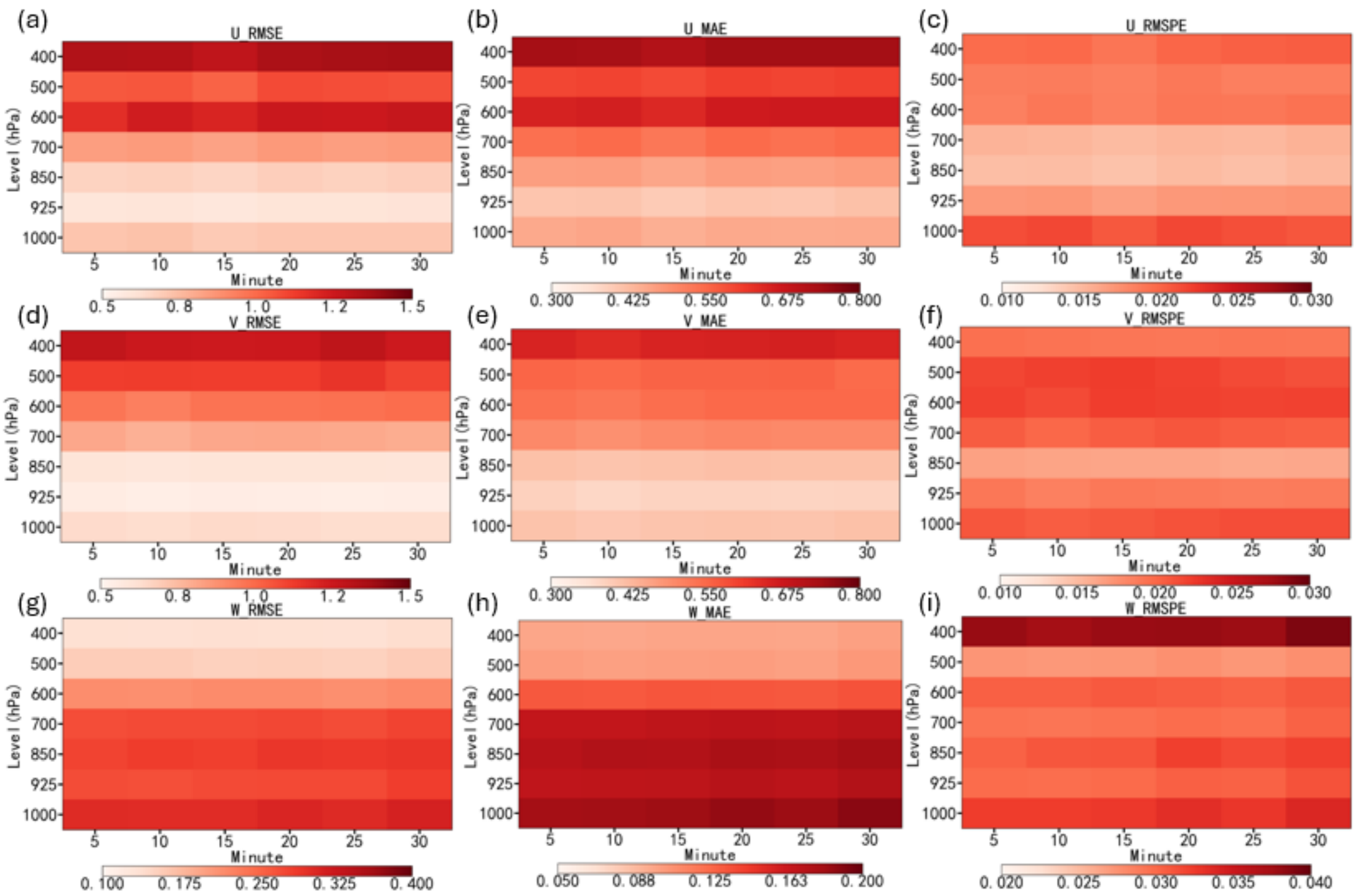

**Fig. 2 | 5G GNSS wind forecast mosaic plots for (a-c)** U-wind, (**d-f**) V-wind and (**g-i**) W-wind predictions at different lead times (5 to 30 minutes at x axis) and heights (1000 to 400 hPa at y axis). From left to right, the first column shows RMSE metrics, the second column shows MAE metrics, and the third column shows RMSPE metrics. The units for RMSE and MAE are m/s, while RMSPE is a dimensionless number ranging between 0-1. These metrics are first calculated at individual stations and then averaged over Zhejiang province of China.

We also present the spatially averaged line chart (i.e., time series) of the 30-minute wind predictions and the true values from the test set, as shown in Fig. 3. It can be observed that for both U and V, whether at 400 hPa, 700 hPa, or 1000 hPa, there is almost no difference between our model's outputs and the true values from the test set, with only minor discrepancies in very few instances. For W, the results are slightly worse than those for U and V, which can be attributed to the fact that vertical velocity is influenced by more factors, not limited to moisture[49]. The predictions at 400 hPa appear to be the poorest, which is consistent with the conclusion drawn from Fig. 2: the geopotential height differences between higher pressure levels are larger, and there is no data output above 400 hPa, making it difficult for the model to capture moisture gradients and W gradients, thus increasing the prediction difficulty. However, overall, our model still demonstrates highly excellent results, with the predicted U, V, and W at all heights closely matching the true values from the test set for a 30-minute lead time.

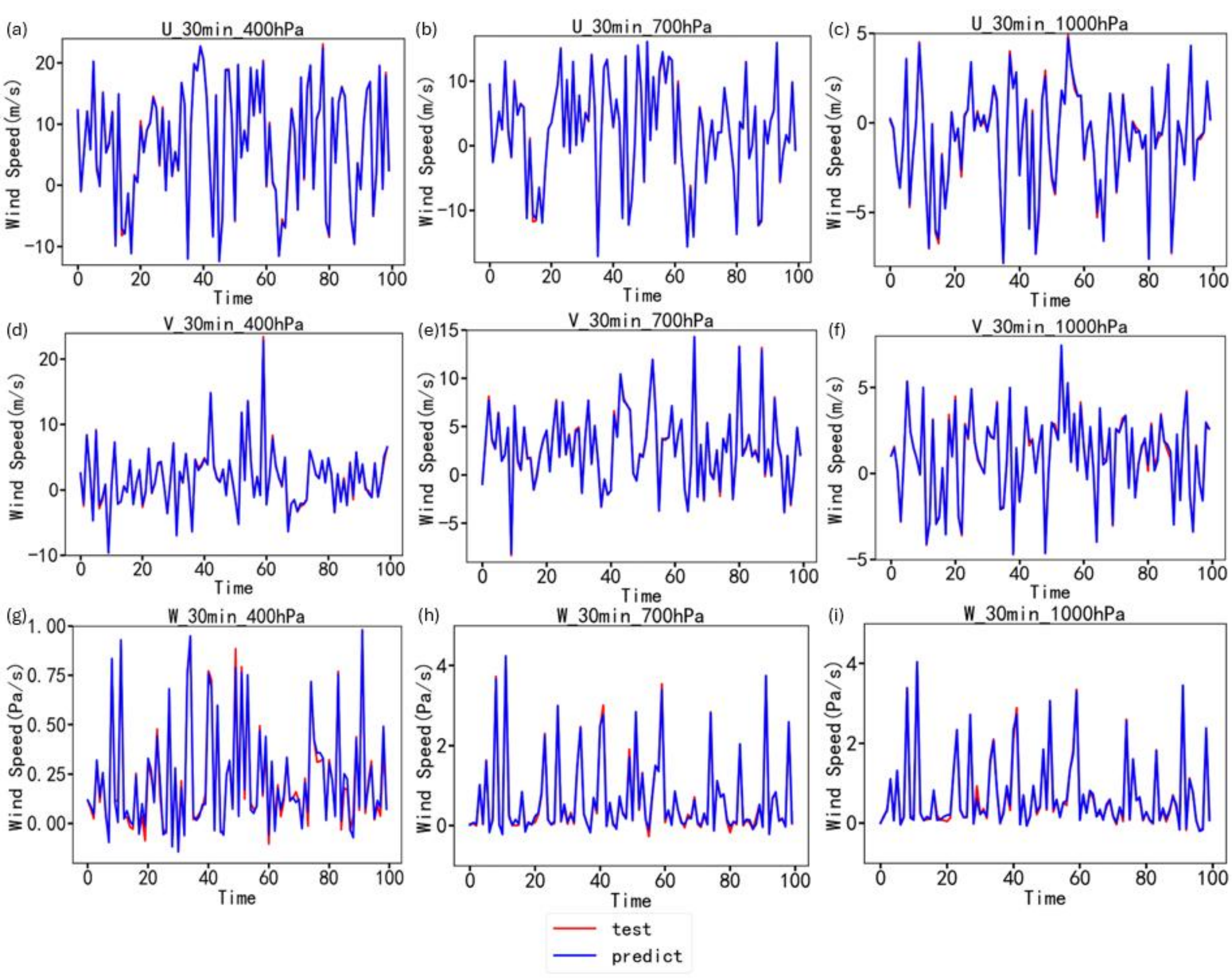

**Fig. 3 | Line charts of spatially average 30-minute-lead wind predictions (blue) and ground-true values** (red) for (**a-c**) U, (**d-f**) V and (**g-i**) W. From left to right: the first column represents the 400 hPa, the second column the 700 hPa, and the third column the 1000 hPa.

All wind speeds are in units of m/s.

### Spatial consistency of G-WindCast performance

To demonstrate spatial consistency of our model performance, we plotted the spatial distribution of multi-level hourly wind predictions at 30-minute lead, the corresponding ground truth at RWP stations, and ERA5 wind reanalysis averaged over the test set (Fig. 4). We note that only the wind predictions on the hour are compared because ERA5 merely provides the wind reanalysis on the hour. It is clear that our model's predictions (first row of Fig. 4) closely match the ground truth (second row of Fig. 4) in both wind speed and direction, whether inland or near the ocean at different vertical levels. Typically, sea surface wind speeds are easier to predict than land surface winds due to the more complex conditions over land, which is why recent 5G GNSS-based wind derivation efforts have primarily focused on sea surface winds[50, 51]. However, our experiments indicate that the underlying surface conditions do not significantly affect our wind derivation, demonstrating the robustness of our model.

Furthermore, we compared the ERA5 multi-level wind reanalysis with the ground-true values in the test set (Figs. 4d-i). We matched the ERA5 to the nearest grid points of RWP stations. We note that unlike our model's 30-minute predictions, the ERA5 reanalysis data is a fusion of numerical models, satellite, radar, and other diverse sources[52], and it is not a forecast result. Even under these conditions, significant discrepancies remain between the ERA5 wind reanalysis and the observed true values. At 400 hPa, while the wind direction differences are relatively small, the wind speed is notably higher than the true values. At 700 hPa, the discrepancies in both wind speed and direction further increase. Near the surface at 1000 hPa, the wind direction difference between ERA5 and the true values approaches 90˚. Overall, the wind direction differences between ERA5 and the true values gradually widen from higher to lower atmospheric levels. As for wind speed, ERA5 consistently overestimates compared to the true values. This further demonstrates the robustness of our model—even its 30-minute forecast results are closer to the observed true values than the contemporaneous ERA5 reanalysis data.

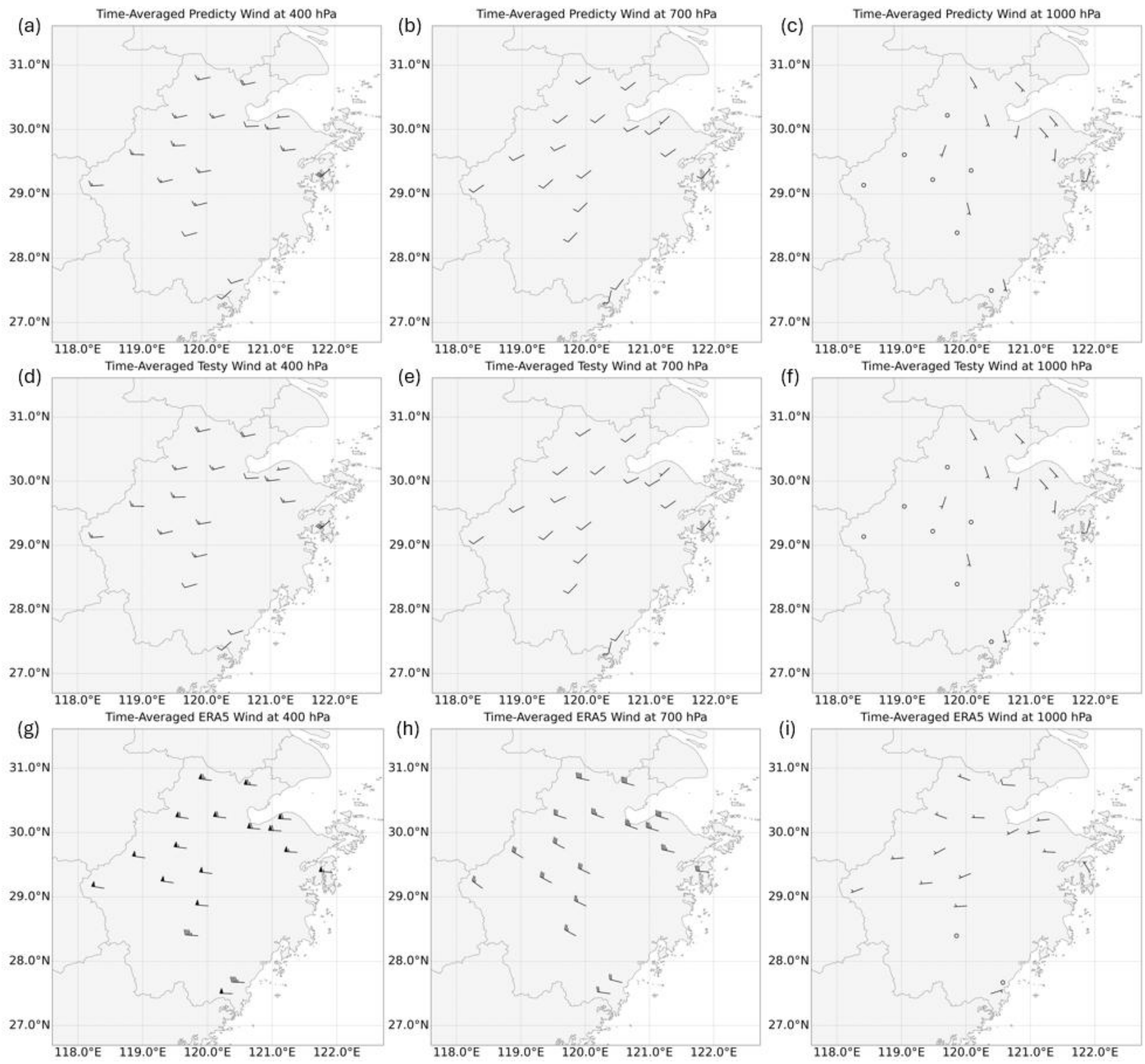

**Fig. 4 | Spatial distributions of multi-level (a-c) 30-minute-lead wind predictions (d-f) ground-true values and (g-i) ERA5 wind reanalysis on the same hours averaged over the test set**; from left to right: first column shows 400 hPa, second column shows 700 hPa, third column shows 1000 hPa. The wind barbs in the figure indicate wind speed and direction. The shaft of the barb points in the direction from which the wind is blowing. Each half barber represents 2.5 m/s, each full barber represents 5 m/s, and each flag represents 25 m/s. For example, a full barb and a half barb indicate a wind speed of 7.5 m/s, while a flag and a full barb represent 30 m/s. Circles denote calm or light winds with speeds less than 2.5 m/s.

**Impacts of 5G GNSS station quantity**

In all the aforementioned experiments in this study, we utilized 1,215 5G GNSS stations in Zhejiang province for training and testing our model. However, in certain scenarios, the number of available 5G GNSS stations may be limited, and the spatial range for wind speed prediction may also be narrower. For instance, an airport may wish to forecast local wind speeds but can only establish 5G GNSS stations within a small surrounding area, while needing to predict multi-level wind speeds at a single point. In this section, we conducted experiments to determine the minimum number of 5G GNSS stations required for wind predictions using 5G GNSS ZTD data. First, based on the hypothetical scenario described, we selected Yiwu Station (Station No. 58557), located near the center of Zhejiang province, with a latitude of 29.3619˚N and longitude of 120.0717˚E. We then logarithmically selected the nearest 20, 45, 100, 230, 520, and 1,215 stations from the available 1,215 5G GNSS stations to Yiwu Station. Using the same model, we conducted six sets of experiments to investigate how the model's performance varied with different numbers of stations. Fig. 5 illustrates the overall performance of the model across these experiments. It reveals that, in general, all metrics gradually decline as the number of 5G GNSS stations decreases. However, this decline is not linear but follows an exponential trend. When using 100 to 1,215 nearby 5G GNSS stations, the metrics show only a slight downward trend as the station count decreases. In contrast, when fewer than 100 stations are used, the decline in all metrics becomes significantly steeper. For U, V, and W wind components, since the RMSE and MAE values for W are notably smaller than those for U and V, the decrease in W's RMSE and MAE metrics appears less pronounced when fewer than 100 stations are used. However, for dimensionless metrics like RMSPE and correlation coefficients, W exhibits a similar decline trend to U and V when station counts drop below 100. The inflection point in Fig. 5 suggests that our model requires at least 100 5G GNSS stations to achieve relatively strong performance in wind speed prediction. Below this threshold, model performance deteriorates sharply.

To provide a more intuitive comparison between using 100 and 1,215 5G GNSS stations, we plotted radar charts of their respective metrics, as shown in Fig. 6. For clarity and uniform scaling, we divided RMSE and MAE by 10. Since higher correlation coefficients (R) are better and range between -1 and 1, we displayed them as 1-R to align with other metrics. Fig. 7 demonstrates that the model using 1,215 stations outperforms the one using 100 stations across U, V, and W wind components. However, the gap is not particularly large, especially for U and V winds, where the results are very close. For W wind, the differences in correlation coefficients and RMSPE between 100 and 1,215 stations are more noticeable but still within an acceptable range.

Finally, we also present a time-series plot (Fig. 7) comparing 30-minute wind prediction versus actual wind at Yiwu Station, using ZTD data from 100 5G GNSS stations. Fig. 7 shows that for U wind, the predictions at 400 hPa and 700 hPa are nearly identical to those using 1,215 stations that are almost identical to the ground-true value and thus not shown. However, at 1,000 hPa, some extreme values are missing. As previously analyzed, surface-level wind speeds are influenced by numerous factors beyond just water vapor, and our model inherently exhibits higher RMSPE for U and V at 1,000 hPa (Fig. 2). Further reducing station counts would only worsen this effect. For V, while performance degradation at mid-to-low levels (700 hPa and 1,000 hPa), especially 1,000 hPa, is less severe than for U, there are still time periods where the model fails to fully capture observations (indicated by blue lines not covering red lines). At 400 hPa, V predictions are nearly indistinguishable from those using 1,215 stations (The figure from station 1215 is not displayed). For W, performance degrades across all pressure levels compared to using 1,215 stations (The figure from station 1215 is not displayed.). At 700 hPa and 1,000 hPa, two extreme values are missed, and other non-extreme predictions are also less accurate. At 400 hPa, the model's performance declines significantly, with most predicted values deviating noticeably from actual observations.

Overall, in this experiment, using only 1/12 of the available stations still yielded excellent performance. These results are encouraging. For the initial hypothetical scenario—local forecasting at an airport—establishing just 100 5G GNSS stations around the airport would suffice to achieve strong wind speed predictions 30 minutes ahead.

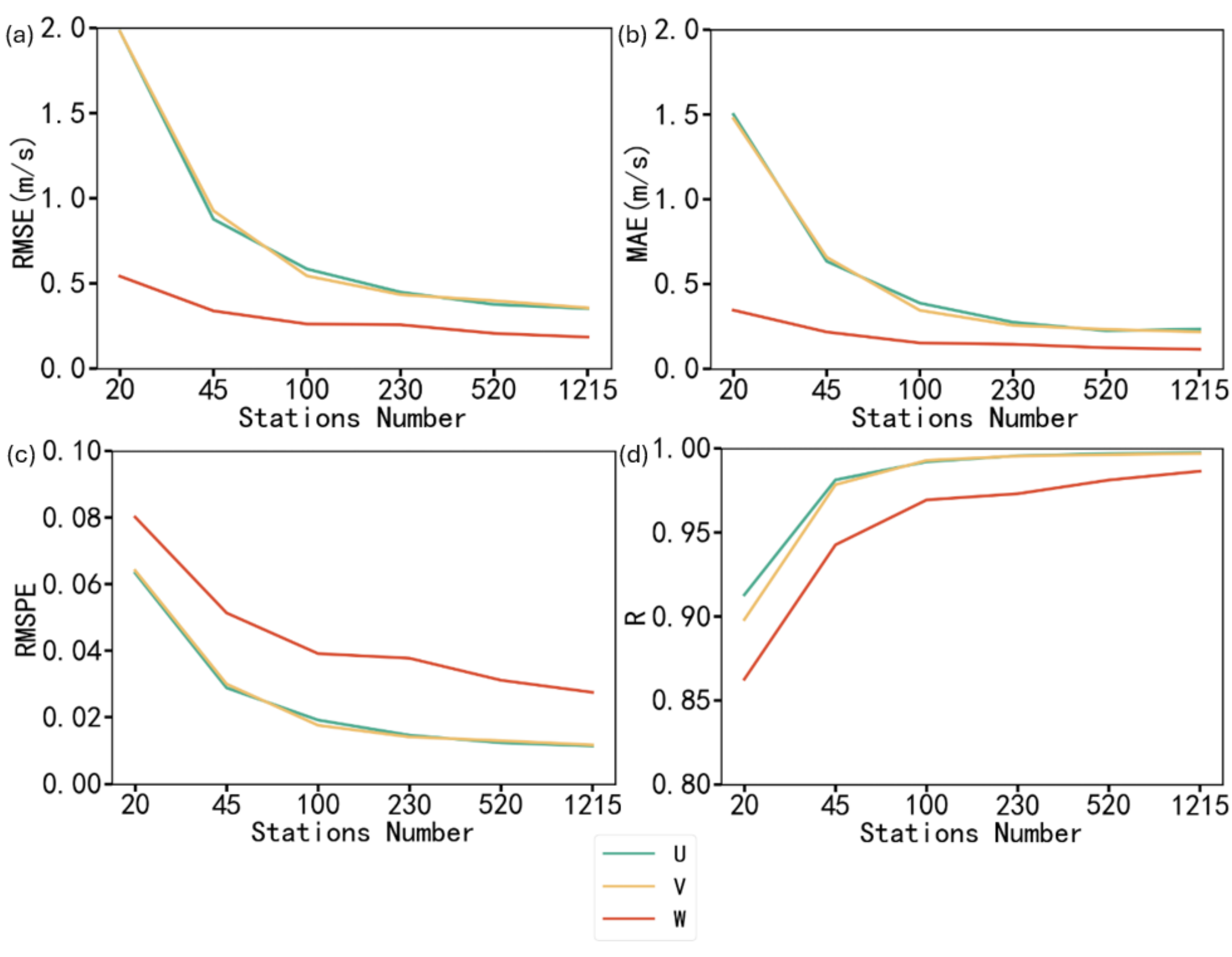

**Fig. 5 | The metrics between 30-minute-lead predictions and true values of wind speed at**

**Yiwu Station in Zhejiang Province using different numbers of 5G GNSS stations** (**a**) RMSE (**b**) MAE (**c**) RMSPE (**d**) correlation coefficient; RMSE and MAE have units of m/s, while RMSPE and correlation coefficient are dimensionless.

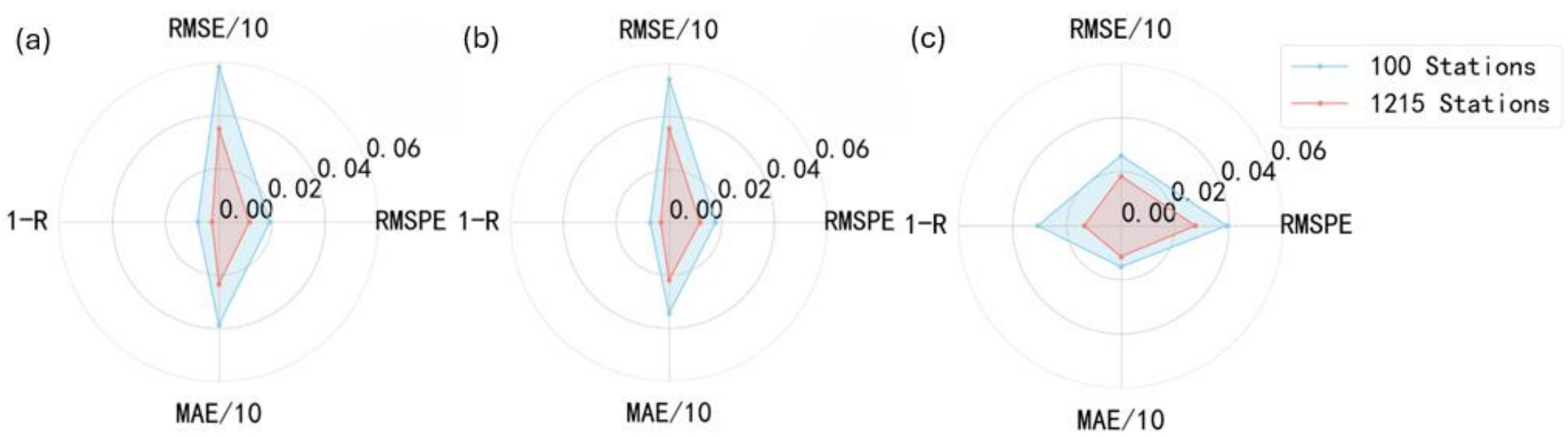

**Fig. 6 | Comparison of RMSE, MAE, RMSPE, and correlation coefficients between 30-minute-lead predictions and true values at Yiwu Station in Zhejiang, using 100 (light blue) and 1215 (light red) 5G GNSS stations** for (**a**) U, (**b**) V, and (**c**) W; RMSE and MAE have units of m/s, while RMSPE and correlation coefficients are dimensionless.

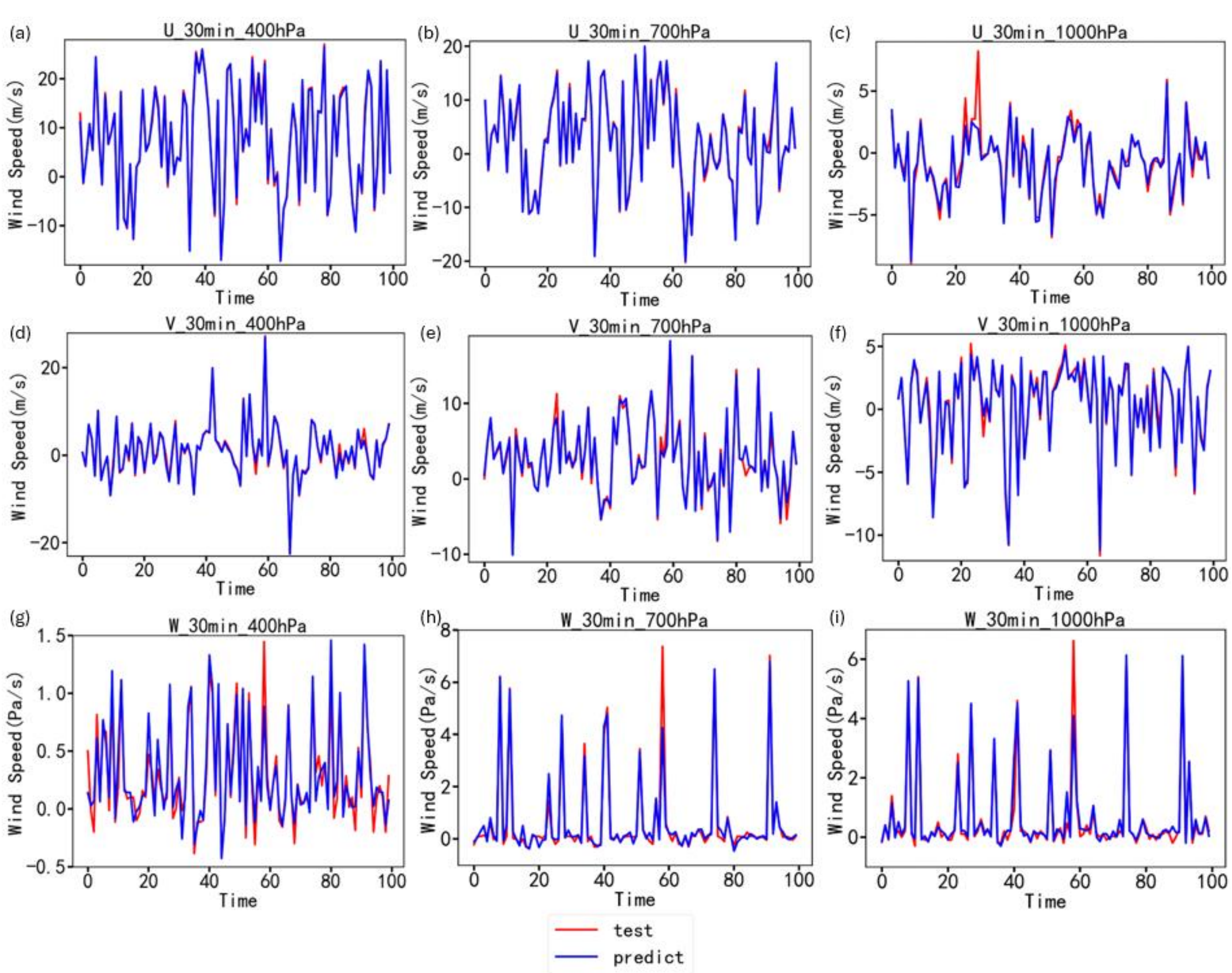

**Fig. 7 | Line charts of spatially-average 30-minute-lead wind predictions (blue) and ground-true values** (red) for (**a-c**) U, (**d-f**) V and (**g-i**) W **at Yiwu Station using 100 5G**

**GNSS stations**; The first column from left represents the 400 hPa, the second column the 700 hPa, and the third column the 1000 hPa; all wind speeds are in units of m/s.

## Discussion

This paper proposes a novel deep learning framework, G-WindCast, which leverages 5G GNSS signal strength variations to predict three-dimensional (U, V, W) atmospheric wind fields in real time. By integrating deep learning models such as the Transformer, we successfully captured the complex nonlinear relationships between GNSS signal variations and wind fields, achieving significant results in short-term (5–30 minutes) wind field prediction tasks.

The main conclusions of this study include:

1. High Accuracy and Robustness: Our model demonstrated stable performance across 5–30-minute predictions for U, V, and W wind components at different pressure levels, with metrics such as RMSE, MAE, and RMSPE showing no significant degradation as forecast lead time increased, and RMSPE consistently below 0.04 (4%), confirming the model's robustness and high accuracy.

2. Superiority Over Reanalysis Data: Compared to the state-of-the-art ERA5 multi-level reanalysis data, our model's predictions—even at 30-minute lead times—were closer to ground-based RWP in both wind speed and direction, especially at lower altitudes where ERA5 showed notable deviations. This further validates the effectiveness and superiority of our model.

3. Broad Applicability: The model's predictions exhibited high spatial consistency with true values in the test set, demonstrating strong agreement in wind speed and direction across both inland and coastal regions. This highlights the model's adaptability to diverse underlying surface conditions.

4. Efficiency and Scalability: The study found that even with a substantial reduction in the number of 5G GNSS stations (e.g., using only ~100 stations, or 1/12 of the total), the model could still provide reasonably accurate 30-minute wind speed predictions for localized areas (e.g., airports). This underscores the method's potential for cost-effective real-world deployment.

Despite these positive outcomes, the study has several limitations. In recent years, weather prediction using large foundation models has proliferated[53-55]. These models leverage extensive meteorological datasets, often comprising over 100 variables, with temporal coverage spanning decades and spatial coverage extending globally. In contrast, our study utilized only three months of 5G GNSS ZTD data, limited to Zhejiang province, which is insufficient for employing large meteorological models in terms of both spatial and temporal scope. Additionally, due to the limited temporal coverage, our dataset was split randomly rather than sequentially to ensure the i.i.d (independent and identically distributed) assumption for the training and test sets as the wind fields have strong seasonality. However, random splitting increases the risk of overfitting. Future work should explore whether acquiring longer-term and broader spatial data could facilitate the use of more complex models (or even large meteorological models). With extended temporal and spatial coverage, sequential dataset splitting could mitigate overfitting and enhance model robustness. Furthermore, this study relied solely on ZTD data for wind field prediction. Recent research has demonstrated the benefits of integrating multi-source meteorological observations (e.g., satellite, radar, and ground-based data) as model inputs[56]. Future efforts could investigate whether combining such diverse data sources could further improve short-term wind field prediction performance.

In summary, this study presents a promising new approach for high spatiotemporal-resolution wind field prediction by combining high-density, high-temporal-resolution ZTD data from 5G GNSS stations with deep learning methods. It opens new possibilities for meteorological forecasting and environmental monitoring.

**Data Availability**

The data used in this study were sourced from the Meteorological Observation Engineering Technology Research Center of the China Meteorological Administration and Huawei Technologies Co., Ltd. These are confidential data, and the authors are not authorized to disclose them.

**Code availability**

Use the Anaconda software (https://www.anaconda.com/download), which includes Jupyter Notebook as the foundational programming software, using Python as the primary programming language. All fundamental deep learning modules are sourced from the open-source library TensorFlow (https://tensorflow.google.cn/). The code used in the article has been uploaded to GitHub (https://github.com/qq492947833/G-WindCast).

# Appendices

# A Deep Learning Model for Wind Field Prediction Using 5G GNSS Signal

## A.1 Transformer

The Transformer was proposed by Vaswani et al. [1] as a natural language processing (NLP) model. It belongs to a variant of RNN [2]. Compared to RNN, the Transformer can process input data in parallel through positional encoding, whereas RNN requires the computation results from the previous time step as input for the next, making parallel computation impossible. The Transformer operates more efficiently than RNN but also consumes more resources.

The module enabling parallel input processing in the Transformer is the Multi-head Attention mechanism. This module multiplies each input data from time steps $x_0$ to $x_t$ with weight matrices $W_{0-t}^{q}$, $W_{0-t}^{k}$ and $W_{0-t}^{v}$ to obtain $q_{0-t}$, $k_{0-t}$, and $v_{0-t}$. Subsequently, the k matrices of each time step are multiplied with the q matrices of other time steps, and the resulting product is then multiplied with the v matrix of that time step, followed by the addition of a bias term b to produce the final output. In this way, variables at each time step undergo matrix operations with those at other time steps, and the final result incorporates relationships between all time steps (including themselves). Additionally, the input $x_t$ at each time step is obtained by adding positional encoding to the original input, ensuring the final result retains sequential information.

The Transformer consists of an encoder and a decoder. The encoder processes input features, while the decoder iteratively generates outputs (using the previous time step's output to derive the next). However, in this study, multiple time steps of multiple predictors correspond to a single output value, so only the Transformer's encoder module is used. This module takes predictors as input, applies positional encoding to each time step of each predictor, and feeds the combined result into the Multi-head Attention mechanism. After processing by the mechanism, the input data is combined with the output via residual connections, followed by normalization. Finally, a fully connected layer performs inference and prediction, with another residual connection between the pre-fully-connected input and the processed result, which is then normalized to produce the final output.

In this study, the Transformer's structure is relatively simple to prevent overfitting—after processing, it directly passes through a fully connected layer for output. The internal fully connected layer has the same number of neurons as the input and also uses the Tanh activation function. The Transformer's network architecture is illustrated in SI Fig. 1.

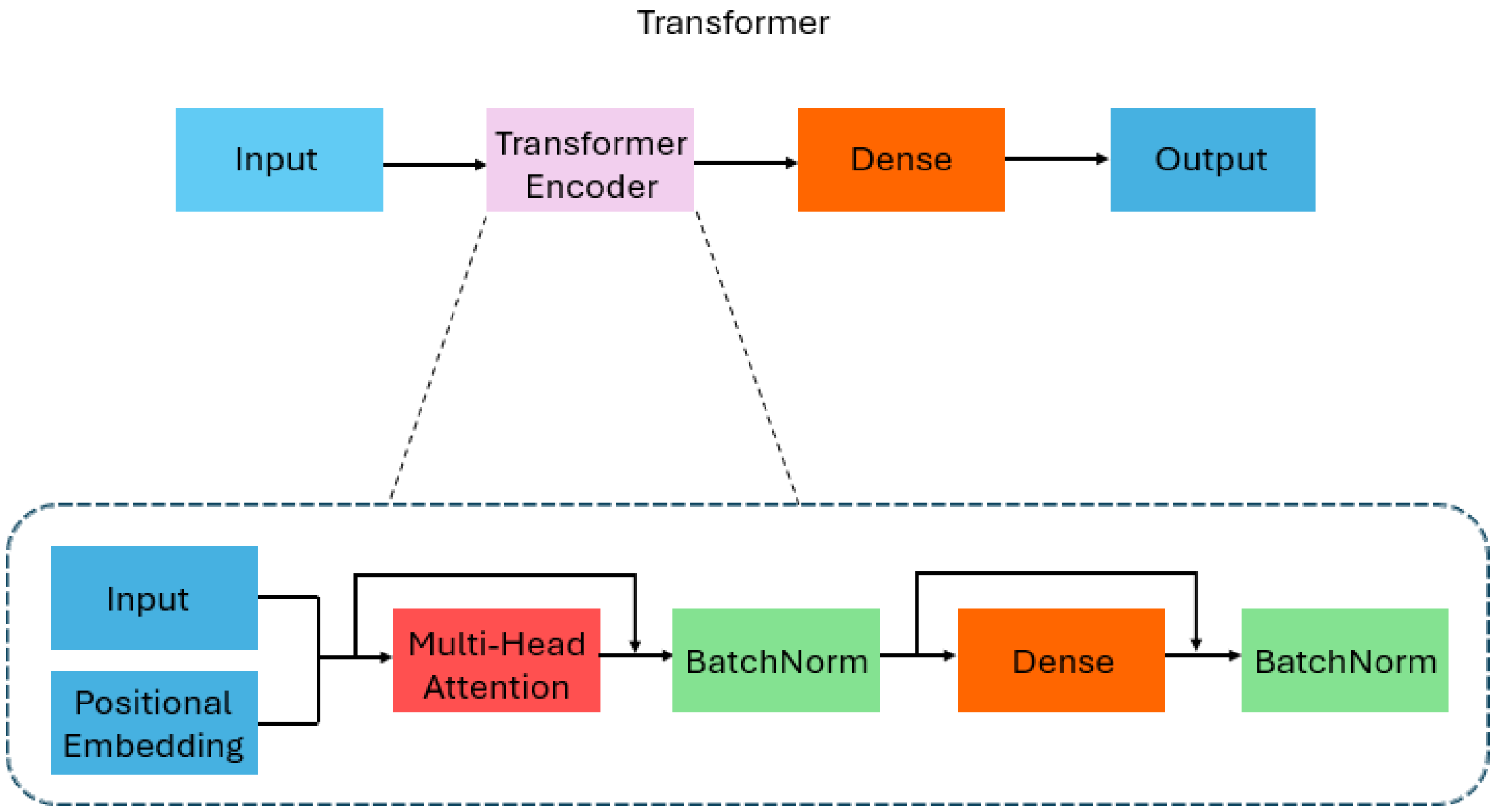

**Supplementary Figure 1:** The Transformer network architecture used in the experiment; where Inputs and Outputs represent the input and output layers; Positional Embedding denotes the positional encoding layer; Multi-Head Attention stands for the multi-head self-attention layer; BatchNorm refers to the normalization layer; Dense represents the fully connected layer; the Transformer Encoder can be reused N times in practical applications.

# B Post-processing

## B.1 CDF Matching

The Cumulative Distribution Function (CDF) matching method is widely used in remote sensing and meteorology [3-5]. For a random variable X, its cumulative distribution function $F_X(x)$ is defined as $F_X(x) = P(X \leq x)$ , representing the probability that the random variable X is less than or equal to a certain value x. Suppose we need to transform the random variable X into another random variable Y, with the target distribution's CDF $F_Y(y)$ known. First, we calculate the CDF $F_X(x)$ of the source random variable X, then convert the values of X into a uniformly distributed random variable: U = $F_X(x)$ ,where U is a uniform random variable on [0,1]. Finally, we transform the uniform random variable U into the target random variable using the inverse function of the target CDF, $F_Y^{-1}(u)$: Y = $F_Y^{-1}(u)$. The resulting Y after these steps represents a sample of the random variable that conforms to the target distribution.

This study employs CDF matching to adjust the mean and variance of the model's output on the test set, bringing them closer to those of the training set. This approach prevents the model's output from being systematically higher or lower than the true values in the test set. When applying the model subsequently, the mean and variance of the training set are still used for CDF matching, eliminating the need for true values from test set samples and avoiding any data leakage issues.

# C Evaluation Metrics

This paper employs the following five commonly used meteorological evaluation criteria: root mean square error (RMSE), percentage root mean square error (RMSPE), mean absolute error (MAE), and correlation coefficient (R).

## C.1 RMSE

The RMSE formula is as follows:

$$RMSE = \sqrt{\frac{1}{mn}\sum_{i=1}^{m}\sum_{j=1}^{n}(\tilde{Y}(i,j) - Y(i,j))^2} \quad (1)$$

where M and N represent the number of latitude grid points and longitude grid points, respectively, $c$ denotes the model predicted values, and Y represents the true values. The RMSE ranges from 0 to +∞, with smaller values indicating better model predictive performance.

## C.2 MAE

The formula for MAE is as follows:

$$MAE = \frac{1}{mn}\sum_{i=1}^{m}\sum_{j=1}^{n} ABS(\tilde{Y}(i,j) - Y(i,j))) \quad (3)$$

where M and N represent the number of latitude grid points and longitude grid points, respectively, $\tilde{Y}$ denotes the model predicted values, Y represents the true values, and ABS stands for taking the absolute value. The MAE ranges from 0 to +∞, with smaller values indicating better model predictive performance.

## C.3 RMSPE

The formula for RMSPE is as follows:

$$RMSPE = \frac{1}{mn} \sum_{i=1}^{m} \sum_{j=1}^{n} \frac{RMSE(i,j)}{MAX(Y(i,j)) - MIN(Y(i,j))} \tag{2}$$

where M and N represent the number of latitude grid points and longitude grid points, RMSE stands for Root Mean Square Error, MAX represents the maximum value, MIN denotes the minimum value, and Y indicates the true value. The RMSPE ranges from 0 to +∞, with smaller values indicating better model predictive performance.

## C.4 Pearson Correlation Coefficient

The formula for R is as follows:

$$R = \frac{1}{mn} \sum_{i=1}^{m} \sum_{j=1}^{n} \frac{\sum (\tilde{Y}(i,j) - \bar{\bar{\tilde{Y}}}(i,j))(Y(i,j) - \bar{Y}(i,j))}{\sqrt{\sum (\tilde{Y}(i,j) - \bar{\bar{\tilde{Y}}}(i,j))^2 (Y(i,j) - \bar{Y}(i,j))^2}} \tag{4}$$

where M and N represent the number of latitude grid points and longitude grid points, respectively, $\tilde{Y}$ denotes the model predicted values, Y represents the true values, $\bar{\bar{\tilde{Y}}}$ denotes the time average of $\tilde{Y}$, $\bar{Y}$ denotes the time average of Y. The range of R is from -1 to 1, where -1 indicates a perfect negative correlation, 1 indicates a perfect positive correlation, and 0 indicates no correlation at all. For model predictions, we want R to be as close to 1 as possible.

# D Training

## D.1 Optimizer

This paper employs the Adam optimizer [6] , a widely used adaptive learning rate optimization algorithm in deep learning, originally proposed by Diederik and Jimmy in 2014. Adam offers the advantage of automatically adjusting the magnitude of parameter updates, enabling faster convergence and better performance when training deep neural networks.

The core idea of Adam is to adaptively adjust the learning rate for each parameter individually, utilizing both the first-order moment (mean) and the second-order moment (uncentered variance) for parameter updates. The update steps of Adam are as follows: 1. Assume the current parameter is $\theta$, and the gradient of the loss function with respect to the parameter is $g_t$ (where t denotes the epoch). First, initialize the first-order moment $m_0 = 0$ and the second-order moment $v_0 = 0$; 2. Update the first and second moments: $m_t = \beta_1 m_{t-1} + (1 - \beta_1) g_t$, $v_t = \beta_2 v_{t-1} + (1 - \beta_2) g_t^2$, where $\beta_1$ and $\beta_2$ are the decay coefficients for the first and second moments, typically set to $\beta_1$=0.9 and $\beta_2$=0.999; 3.Perform bias correction: Since $m_0$ and $v_0$ are initialized to zero, there is an initial bias in the early stages of training. Thus, compute the corrected moments: $\widehat{m}_t = \frac{m_t}{1-\beta_1^t}$, $\hat{v}_t = \frac{v_t}{1-\beta_2^t}$; 4.Update the parameter:

$\theta_{t+1} = \theta_t - \alpha \frac{\widehat{m}_t}{\sqrt{\hat{v}_t + \epsilon}}$, where $\alpha$ is the learning rate, and $\epsilon$ is a small constant to prevent division by zero. The term $\frac{\widehat{m}_t}{\sqrt{\hat{v}_t + \epsilon}}$ effectively weights the learning rate. In the early stages of training, $\hat{v}_t$ is small, resulting in a larger learning rate, while in later stages, $\hat{v}_t$ increases, leading to a smaller learning rate.

Through this adaptive adjustment, Adam enables larger parameter updates early in training, allowing rapid loss reduction, while later updates become smaller, facilitating convergence toward the global optimum.

## D.2 Loss Function

This paper employs the Mean Squared Error (MSE) as the loss function. MSE assigns greater weight to larger errors through squared differences, which proves particularly advantageous for forecasting extreme wind speeds [7] . Extreme wind speeds represent the most critical factor threatening aviation and related fields [8] . Therefore, all models in this study utilize MSE as the loss function.

## D.3 Learning Rate

The learning rate determines the magnitude of parameter adjustments during each optimization step of the network. A higher learning rate results in larger parameter updates, and vice versa. An excessively large learning rate may prevent the network from converging, while an overly small learning rate requires more epochs for the network to approach the global optimum. In this paper, all models were trained for 5,000 epochs, with a learning rate of 0.00001 proving to be most appropriate [9].

## D.4 Early Stopping

Early stopping is a commonly used strategy in machine learning and deep learning to prevent overfitting and reduce resource consumption during network training [10] . The principle of early stopping is straightforward: first, set a tolerance value P. During network training, the validation set loss function is monitored in real-time for each epoch. If the validation set loss function does not decrease within P epochs after reaching an optimal epoch, the training is halted. The model parameters corresponding to the optimal loss function are then recorded and reloaded to ensure the best validation results. A tolerance value that is too high may lead to unnecessary computational resource consumption, while a value that is too low may prevent the network from finding the global optimal parameters. In this study, the number of epochs is set to 5000, and the tolerance P is set to 1000. This ensures that excessive computational resources are not wasted while allowing the network sufficient patience to find the globally optimal parameters.

# E Data

## E.1 5G GNSS ZTD Data

The 5G GNSS ZTD data used in this study were obtained from GNSS stations operated by Huawei and China Mobile. This dataset is two-dimensional (time, station), and the authors do not have the authority to disclose it. The time periods of data usage and the specific number of stations are shown in SI Table 1. Within each time period, the data are continuous with a resolution of 5 minutes. The number of stations remains fixed within each time period but varies across different periods. Based on the 5G GNSS ZTD data used, we divided the study into two experiments: Experiment 1 prioritized temporal samples, meaning all available time periods were used. For station selection, only the common stations shared across all time periods were retained, resulting in 21,701 time steps and 566 stations. Experiment 2 prioritized station samples, meaning only time periods with at least 1,000 stations were selected. From these, common stations were identified, yielding 17,594 time steps and 1,215 stations. The spatial distribution of stations for both experiments is illustrated in SI Fig. 2. Since the dataset contains missing values, we applied bilinear interpolation to ensure that the final input data fed into the model were complete and free of gaps.

**Supplementary Table 1：** The distribution of time periods, number of stations, and experimental information for the 5G GNSS ZTD data used in the study.

| Time Period | Number of Stations | Using Experiments |
|---|---|---|
| 2025-05-07 05:30:00-2025-05-15 00:35:00 | 885 | Experiment 1 |
| 2025-05-15 13:10:00- | 1392 | Experiment 1 |

| 2025-05-21 23:55:00 | | Experiment 2 |
| --- | --- | --- |
| 2025-05-22 07:30:00-<br>2025-05-30 08:50:00 | 935 | Experiment 1 |
| 2025-05-29 18:00:00-<br>2025-06-06 01:35:00 | 1545 | Experiment 1<br>Experiment 2 |
| 2025-06-07 22:40:00-<br>2025-06-15 07:25:00 | 1545 | Experiment 1<br>Experiment 2 |
| 2025-06-16 10:55:00-<br>2025-06-18 16:00:00 | 1502 | Experiment 1<br>Experiment 2 |
| 2025-06-24 05:45:00-<br>2025-07-01 07:10:00 | 1542 | Experiment 1<br>Experiment 2 |
| 2025-07-01 11:55:00-<br>2025-07-07 01:00:00 | 1440 | Experiment 1<br>Experiment 2 |
| 2025-07-07 07:40:00-<br>2025-07-14 06:15:00 | 1440 | Experiment 1<br>Experiment 2 |
| 2025-07-13 13:35:00-<br>2025-07-21 00:45:00 | 1428 | Experiment 1<br>Experiment 2 |

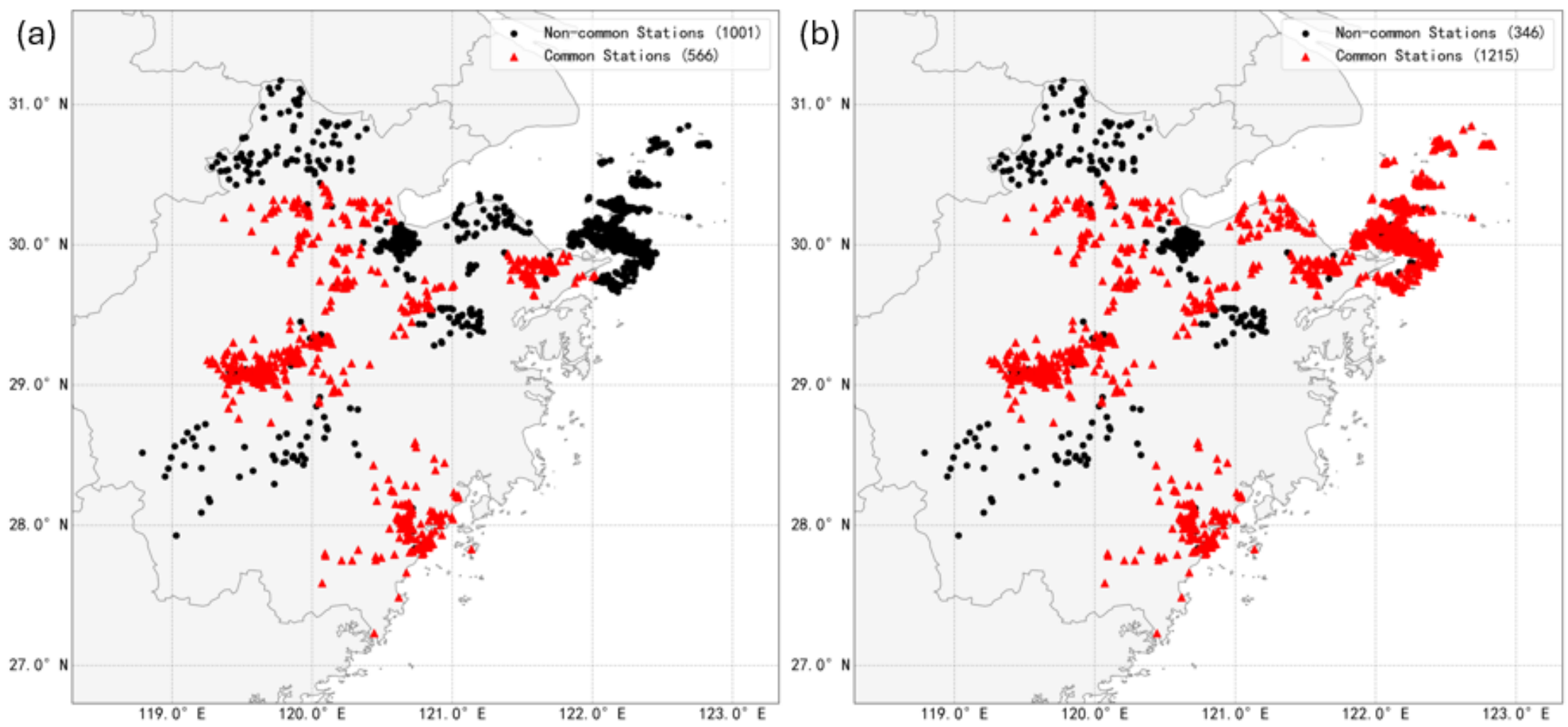

**Supplementary Figure 2:** Spatial distribution of 5G GNSS ZTD data stations in the experiment. (a) Spatial distribution of stations in Experiment 1 (b) Spatial distribution of stations in Experiment 2; red stations in the figure represent those used in the experiment, while black stations represent those not used.

## E.2 Wind Field Data

The wind field data used in this study was sourced from the Meteorological

Observation Center of the China Meteorological Administration. This dataset is three-dimensional (time, altitude, station), and the authors do not have the authority to publicly disclose it. The temporal resolution of the data is 6 minutes, which we interpolated to 5 minutes in our research to align with the 5G GNSS ZTD data's temporal resolution. Within the study area, there are a total of 18 stations, with their spatial distribution shown in SI Fig.3. For the experiments, we selected 7 altitude levels (110m, 760m, 1460m, 3010m, 4200m, 5570m, 7160m) and also attempted to interpolate the altitude-level data to 7 pressure levels (1000hPa, 925hPa, 850hPa, 700hPa, 600hPa, 500hPa, 400hPa) using empirical formulas:

$$P = P_0 * e^{\frac{H}{H_{scale}}} \qquad (5)$$

Where P is the interpolated pressure, $P_0$ is the standard sea-level pressure, taken as 1013.25 hPa in this study, H is the corresponding geopotential height, and $H_{scale}$ is the scale height, taken as 8000 m.

We will compare the effects of both in the experiment. The original variables of this data only include horizontal wind speed, horizontal wind direction, and vertical wind speed. Through the formula:

$$u = -V * Sin(\theta) \qquad (6)$$

$$v = -V * Cos(\theta) \qquad (7)$$

Here, u represents the zonal wind, v the meridional wind, V the horizontal wind speed, and $\theta$ the horizontal wind direction.
The dataset also contains missing values, which we similarly fill using bilinear interpolation to ensure the model input data is free of missing values.

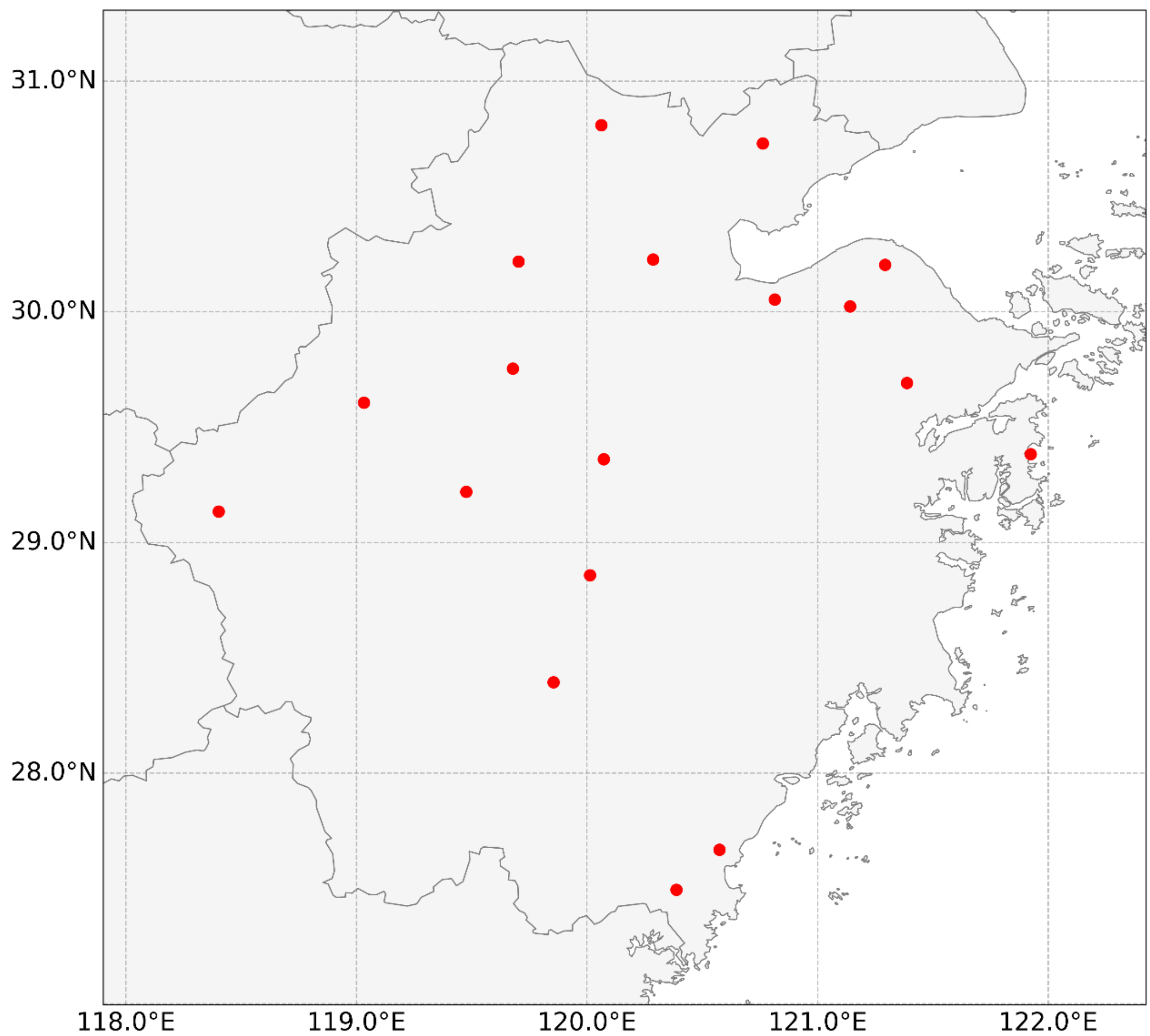

**Supplementary Figure 3:** Spatial distribution map of wind speed data stations in the experiment; the red dotted positions represent the specific spatial locations of the stations.

## E.3 ERA5 Hourly Data on Pressure Levels from 1940 to Present

In this study, to demonstrate the superiority of our model, we employed the currently advanced reanalysis dataset, the ERA5 multi-level reanalysis data [11] . We utilized the U, V, and W variables at 400hPa, 500hPa, 600hPa, 700hPa, 850hPa, 925hPa, and 1000hPa levels. The temporal scope aligns with that of SI Table 1; spatially, we extracted the latitude and longitude of the stations from SI Fig.3 and used the grid points from the ERA5 multi-level reanalysis data closest to each station as the station's data in the reanalysis. The ERA5 data has a temporal resolution of 1 hour, with no 5-minute resolution data available. To highlight the model's superiority, we directly used ERA5 data contemporaneous with the actual observations as the ERA5 data results.